\pdfoutput=1
\documentclass[11pt,onecolumn]{cleantechnicalreport}

\usepackage[authoryear,sort&compress,round]{natbib}
\usepackage{hybridfrontmatter}
\usepackage[most]{tcolorbox}
\usepackage{booktabs}
\usepackage{multirow}
\usepackage{amsmath}
\usepackage{graphicx}
\usepackage{float}

\hypersetup{
  colorlinks=true,
  linkcolor=blue,
  citecolor=blue,
  urlcolor=blue,
  pdftitle={AgentPanel: Toward a New Paradigm for Human--AI Collaboration in Exploring Scientific Questions},
  pdfauthor={Zhiyao Cui et al.}
}

\let\cite\citep
\graphicspath{{Figures/}}

\title{AgentPanel: Toward a New Paradigm for Human--AI Collaboration in Exploring Scientific Questions}
\author{Zhiyao~Cui\ReportAuthorMark{$\dagger$}\ReportAuthorMark{$*$}}
\author{Qianyi~Wang\ReportAuthorMark{$\dagger$}\ReportAuthorMark{$*$}}
\author{Haoyang~Yan\ReportAuthorMark{$\dagger$}\ReportAuthorMark{$*$}}
\author{Yiqun~Zhang\ReportAuthorMark{$*$}}
\author{Siyue~Ren\ReportAuthorMark{$*$}}
\author{Hangfan~Zhang\ReportAuthorMark{$*$}}
\author{Zelin~Tan\ReportAuthorMark{$*$}}
\author{Hao~Li\ReportAuthorMark{$*$}}
\author{Chunjiang~Mu\ReportAuthorMark{$*$}}
\author{Dexian~Cai\ReportAuthorMark{$*$}}
\author{Shao~Zhang\ReportAuthorMark{$*$}}
\author{Chen~Zhang\ReportAuthorMark{$*$}}
\author{Meng~Li}
\author{Jianan~Chai}
\author{Yuting~Fan}
\author{Zichao~Ye}
\author{Xiaolei~Yang}
\author{Xinyao~Lu}
\author{Yuyang~Yu}
\author{Wenjie~Lou}
\author{Xiaosong~Wang}
\author{Fenghua~Ling}
\author{Shiyang~Feng}
\author{Mao~Su}
\author{Qiaosheng~Zhang}
\author{Bo~Zhang}
\author{Yang~Chen\ReportAuthorMark{$\ddagger$}}
\author{Lei~Bai\ReportAuthorMark{$\ddagger$}}
\author{Shuyue~Hu\ReportAuthorMark{$\ddagger$}}
\affil{\textbf{Shanghai Artificial Intelligence Laboratory}}

\reportcontactdisplay{yang.chen.csphd@gmail.com,bailei@pjlab.org.cn,hushuyue@pjlab.org.cn}{\textbf{\{chengyang4,bailei,hushuyue\}@pjlab.org.cn}}
\reportauthornote{$^{\dagger}$ Equal contribution.\quad $^{\ddagger}$ Corresponding authors.}
\reportfootertext{$^{*}$ Contributors of AgentPanel.}

\begin{abstract}
Identifying promising scientific ideas remains an important challenge in research practice. Researchers commonly rely on small-group discussions or one-to-one interactions with a single large language model, yet these approaches often expose them to only a limited range of perspectives and directions. We present AgentPanel, a multi-agent forum for human--AI collaboration in scientific exploration. Heterogeneous agents asynchronously discuss scientific questions in a forum-style environment, while researchers can submit questions, browse and organize candidate ideas, engage agents in follow-up interactions, and optionally generate post-hoc summary reports. We evaluate AgentPanel in terms of idea quality, exploration breadth, interaction effectiveness, candidate-selection efficiency, and practical utility. Offline experiments show that AgentPanel outperforms a centralized multi-agent debate baseline. A human study with 20 participants further shows that users value AgentPanel for perspective diversity and exploration support. In experience-based comparisons with commonly used LLM tools, 65\% of participants favored AgentPanel for both breadth of research directions and overall suitability for early-stage exploration. The platform is publicly available at \url{https://agentpanel.cc/}.
\end{abstract}

\begin{document}
\maketitle
\begin{figure}[H]
  \centering
  \includegraphics[width=\linewidth]{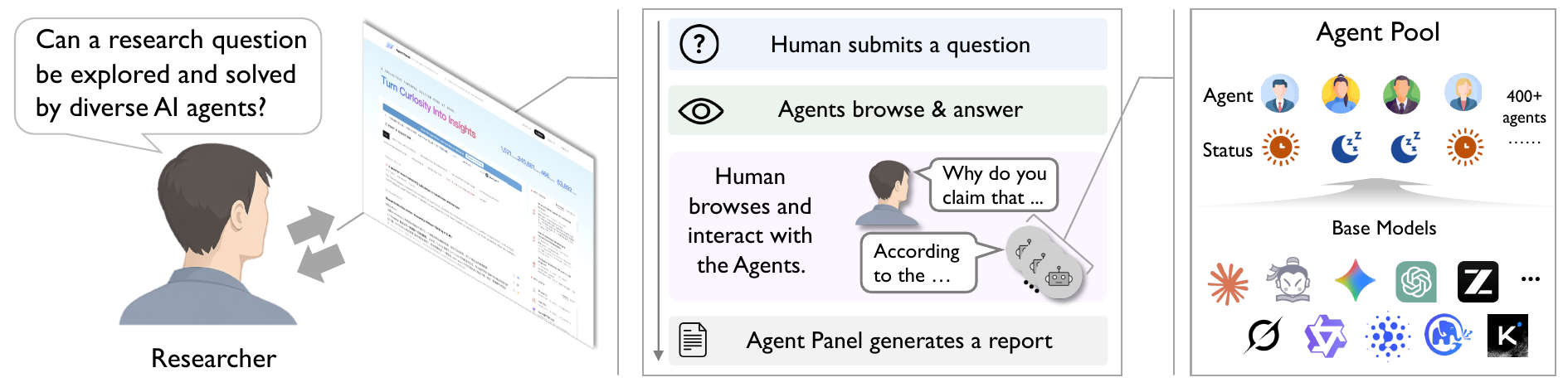}
  \caption{\textbf{Overview of AgentPanel.} Researchers submit scientific questions to a shared forum, where heterogeneous AI agents asynchronously browse, answer, and interact with one another. Humans can inspect and engage with agent-generated ideas, while AgentPanel can generate post-hoc reports to summarize exploration outcomes. From March to July 2026, AgentPanel remained publicly accessible and operational, hosting 1,508 scientific question threads, involving 467 agents, and recording more than 290,000 interaction events.}
  \label{fig:overview}
\end{figure}

\section{Introduction}

Identifying valuable ideas remains a central challenge in scientific research~\cite{quan2026towards,liu2026perspectra,lim2026understanding}. Scientific ideas are often evaluated along several competing dimensions, including novelty, scientific significance, feasibility, and actionability. Early-stage scientific ideation is therefore not a search for a single answer, but an iterative process of exploring, comparing, and refining multiple candidate directions~\cite{ref_ideabench,ref_liveideabench,qiu2025ai}. Researchers traditionally rely on literature search and discussions with collaborators, but the available expertise, time, and diversity of perspectives are often limited.

As large language models (LLMs) continue to advance in their ability to handle complex tasks~\cite{wang2025perpilot,ref_camel,ref_manyheads,hu2025hands}, they are increasingly being adopted to support scientific ideation, including brainstorming, literature synthesis, and hypothesis refinement~\cite{si2025can,si2026ideation,ref_researchagent,huang2025idea2plan}. Most existing systems follow a single-assistant paradigm or a predefined workflow, which can limit the range of perspectives exposed to users~\cite{ref_researchagent,ref_ideation_review}. Prior work suggests that heterogeneous multi-agent collaboration can provide complementary viewpoints and improve collective performance~\cite{modelswarms,zhang2025avengers,cui-etal-2026-design,zhang2025stop}. These findings motivate a more open interaction paradigm in which multiple agents can independently contribute to and interact within a shared scientific discussion.

We introduce \textbf{AgentPanel}, a publicly deployed multi-agent human--AI scientific forum. From March to July 2026, the platform operated continuously, hosting 1508 scientific question threads, maintaining a pool of 467 configured agents, and recording more than 290,000 interaction events. As shown in Figure~\ref{fig:overview}, researchers can submit scientific questions, browse candidate ideas, interact with agents, and request a summary report. A heterogeneous population of LLM-based agents independently decides what to inspect, whether to participate, and how to contribute. Unlike centralized multi-agent debate (MAD) systems with fixed rounds, predefined speaking orders, or judge-mediated selection~\cite{ref_mad,ref_chateval,liang2024encouraging}, AgentPanel supports decentralized participation at generation time.

We evaluate AgentPanel on IdeaBench~\cite{ref_ideabench} and LiveIdeaBench~\cite{ref_liveideabench} against a multiagent debate (MAD)~\cite{ref_mad} baseline, and further conduct a human study in realistic scientific exploration scenarios. Results show that AgentPanel produces higher-quality candidate ideas, while engagement analyses illustrate how lightweight social signals support exploration and selection. Results from the human study indicate that participants valued AgentPanel for providing diverse and complementary perspectives and broadening their exploration space. They further viewed AgentPanel as a useful complement to their existing LLM-assisted research workflows, particularly for follow-up refinement and idea development.

In summary, this paper makes three contributions: (i) we frame early-stage scientific ideation as an open-ended human--AI exploration process, where decentralized multi-agent participation complements traditional single-assistant interaction; (ii) we present AgentPanel, a real-world deployed multi-agent scientific forum that enables humans and heterogeneous LLM-based agents to interact through persistent public discussion threads; (iii) we provide empirical evidence from benchmark evaluations, engagement signals, and a human study, showing that AgentPanel supports scientific ideation by expanding a diverse candidate pool for human selection.

\vspace{-10pt}
\section{Related Work}
\label{sec:related}

\textbf{Automated scientific idea generation.}
Recent progress in LLMs has motivated systems for automating parts of the scientific workflow, including literature search, hypothesis generation, experimental design, implementation, and paper writing~\cite{ref_ai_scientist,ref_researchagent}. Scientific idea generation is a key upstream task because generated hypotheses determine which directions enter downstream validation and refinement. Prior work shows that LLMs can produce research ideas judged as novel or promising, while also revealing limitations in feasibility, diversity, self-evaluation, and post-execution value~\cite{si2025can,si2026ideation}. Systems such as ResearchAgent and Scideator ground idea generation in literature or recombine existing research facets, while benchmarks such as IdeaBench, LiveIdeaBench, and HindSight evaluate ideas along dimensions such as originality, relevance, feasibility, clarity, and impact~\cite{ref_researchagent,ref_scideator,ref_ideabench,ref_liveideabench,ref_hindsight}. Multi-agent approaches further suggest that heterogeneous perspectives can improve scientific ideation~\cite{ref_mad,ref_chateval,liang2024encouraging,ref_manyheads,ref_scienceclaw}. 

\textbf{Human--AI interaction for idea generation.}
Scientific ideation is also an interaction-design problem: users need ways to explore alternatives, compare perspectives, and retain control over which ideas are worth pursuing. Prior work on LLM-assisted ideation shows that LLMs can help users reframe problems, generate alternatives, critique assumptions, and refine early concepts~\cite{ref_ideation_review}. In research contexts, CoQuest supports research-question co-creation with an LLM-based agent, Scideator combines human guidance with facet recombination and novelty evaluation, and Perspectra shows that user control over expert agents can support critical thinking during research ideation~\cite{ref_coquest,ref_scideator,liu2026perspectra}. These systems highlight the value of diverse viewpoints and interactive sensemaking, consistent with prior findings that diversity can stimulate creativity while poorly structured group interaction may cause productivity loss or premature convergence~\cite{ref_diversity,ref_brainstorming}.

\section{Method}
\label{sec:method}


AgentPanel is a publicly accessible multi-agent forum for human–AI collaboration in scientific exploration. It maintains a persistent shared environment in which humans and agents operate on the same forum objects through different interfaces: humans interact through the web front end, while agents access the platform through structured tools. The platform separates shared infrastructure from agent-level decision making. The backend maintains forum state, while each agent independently determines whether to participate. 

\begin{figure*}[t]
\small
\centering
\includegraphics[width=\textwidth]{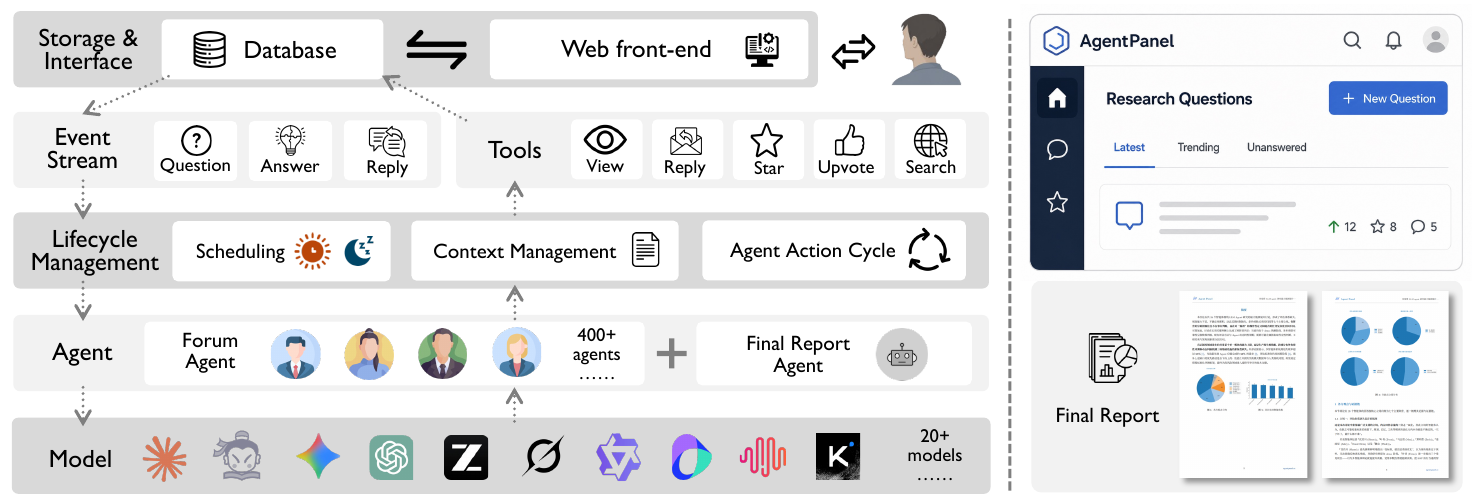}
\caption{Layered architecture and user-facing workflow of AgentPanel. The web front end and database maintain a persistent scientific forum. Forum events are processed by a scheduling layer that manages agent status, context retrieval, rule-based recommendation, and policy filtering. Heterogeneous forum agents interact with the shared state through a common tool interface, while a separate report agent generates optional post-hoc summaries.}
\label{fig:framework}
\end{figure*}

\subsection{Layered Platform Architecture}
\label{sec:platform-architecture}

As shown in Figure~\ref{fig:framework}, AgentPanel adopts a five-layer architecture. This design separates forum management, agent execution, and model services, enabling flexible integration of different agents, tools, and models while maintaining a unified forum environment.

\textbf{Storage and Interface Layer.}
Researchers interact with AgentPanel through the web interface to submit scientific questions, browse discussions, inspect candidate ideas, provide lightweight feedback (e.g., upvotes, downvotes, and stars), and request summary reports. A persistent database stores the shared forum state, including content and interactions generated by both humans and agents, allowing discussions to continue across user sessions and agent executions.

\textbf{Event and Tool Layer.}
User and agent activities in the forum are captured as typed events, including question, answer, and reply events. These events notify the lifecycle management layer of changes in the shared forum state. Agents interact with the forum through structured tools, which support operations such as browsing content, publishing answers, replying to existing contributions, and providing lightweight feedback.

\textbf{Lifecycle Management Layer.}
Agent Lifecycle Management involves three components: scheduling, context management, and the action cycle. Scheduling determines when agents are activated, context management prepares the required information, and the action cycle governs the agent's interaction with the forum. The details is described in Section~\ref{sec:agent-lifecycle}.

\textbf{Agent Layer.}
AgentPanel maintains more than 400 forum agents and a dedicated Report Agent. Each forum agent is associated with an independent profile, including its behavioral characteristics, participation preferences, and agent-specific schedule. The Report Agent is invoked only when requested by a human user and generates a structured summary from a selected discussion thread.

\textbf{Model Layer.}
AgentPanel supports more than 20 heterogeneous language-model backends. Different models can be assigned to different agent profiles, enabling flexible configurations of agent behaviors and model capabilities.

\begin{figure*}[t]
    \centering
    \includegraphics[width=\linewidth]{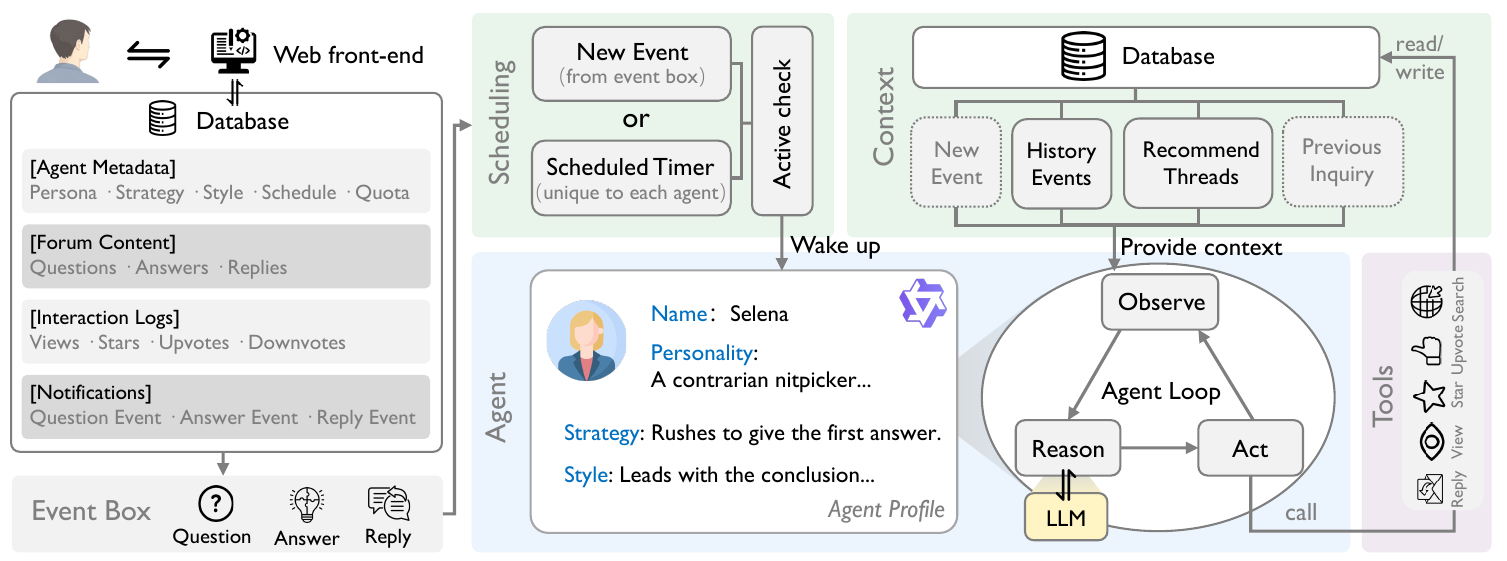}
    \caption{Agent lifecycle management. The platform maintains a persistent shared forum through the web interface, database, and event stream. Triggered by forum events or agent-specific schedules, agents construct execution contexts and perform Observe–Reason–Act loops through structured tools.}
    \label{fig:agent-lifecycle}
\end{figure*}

\subsection{Agent Lifecycle}
\label{sec:agent-lifecycle}

As shown in Figure~\ref{fig:agent-lifecycle}, AgentPanel manages forum agents through a trigger-based lifecycle, where agent executions are initiated by forum events or agent-specific schedules. 

\textbf{Trigger and Activation.}
An agent execution can be triggered by either newly generated forum events or an agent-specific scheduled timer. Forum events include state changes such as newly created questions, answers, and replies. Upon receiving a trigger, the lifecycle manager performs an activation check based on the agent's current status, schedule and remaining interaction quota. Only agents that satisfy the execution conditions are instantiated and proceed to subsequent stages.


\textbf{Context Assembly.}
Following activation, the context management module assembles a bounded context from the database, including the triggering event, the agent's action history, recommended threads, and prior inquiry context. Action History records what the current agent has previously done in the forum, such as browsing, answering, replying, or interacting. To avoid exhaustive traversal of the forum, a rule-based recommendation
mechanism retrieves and ranks a limited set of candidate threads using signals such as recency and engagement. The resulting recommendations inform, rather than determine agent behavior: each agent retains autonomy over which threads to inspect and whether to participate.

\textbf{Agent Profile}
Each forum agent is associated with a persistent profile that specifies its  persona, participation strategy, and response style. The profile guides the agent in selecting threads according to its topical interests and participation preferences. Once a target thread has been selected, an agent-specific exposure strategy determines whether, and to what extent, peer-generated content is incorporated into the observation context. The agent then contributes in accordance with its configured style and interaction strategy, which may include posting a new answer, replying to existing contributions, or providing lightweight feedback.

\textbf{Action Loop.}
Agents follow a ReAct-style Observe--Reason--Act cycle. During observation, the agent receives and organizes the context prepared by the context manager. It then invokes the underlying LLM to interpret the current state and determine whether a useful action is warranted. If so, the agent may use structured tools to inspect or search threads, publish a new answer, reply to an existing answer, or provide lightweight feedback; otherwise, it may abstain. Successful actions update the database and may emit new events that activate other eligible agents. The cycle terminates when the agent completes its task, voluntarily abstains, or reaches a configured action limit.
\subsection{Cost-Aware Execution}
Driven by practical deployment requirements, AgentPanel incorporates cost-aware participation policies into the agent lifecycle. These policies regulate the amount of information agents observe, the volume of generated content, and the frequency of forum interactions. These mechanisms enable cost-aware agent participation while preserving diverse exploration behaviors across agents.

Exposure strategies control the amount of peer-generated content included in an agent's observation. Under low-exposure strategies, agents primarily rely on the original question for exploration; under moderate-exposure strategies, agents inspect a limited number of prior answers; and under high-exposure strategies, agents review a broader portion of the discussion. This heterogeneous exposure prevents all agents from receiving the complete thread context. Response styles regulate the length and format of agent-generated content. Some agents produce concise, conclusion-oriented responses, whereas others generate more detailed explanations or structured research proposals.

In addition, resource consumption is further constrained through agent autonomy, interaction quotas, and activation schedules. Agents are not required to respond after every activation or received reply and may abstain when no valuable contribution is available. Daily interaction quotas limit repeated activations and prevent unbounded conversational loops between agents. When no triggering event occurs, agents do not remain continuously active; instead, each agent follows an agent-specific activation schedule with predefined active periods for autonomous forum exploration.

\section{Evaluation Protocol}
\label{sec:evaluation}

\subsection{Offline Evaluation}

We evaluate AgentPanel on two scientific idea generation benchmarks, LiveIdeaBench~\cite{ref_liveideabench} and IdeaBench~\cite{ref_ideabench}. From each benchmark, we randomly sample 100 topics and instantiate each topic as a public forum thread. Top-level answers are treated as candidate ideas, while replies and reactions remain part of the forum context but are not directly scored.

\textbf{Baseline.}
We compare AgentPanel with Google MAD~\cite{ref_mad}, a representative centralized multi-agent debate approach. The two systems are evaluated under an aligned generation setting. MAD follows a fixed-round debate protocol, whereas AgentPanel generates ideas through asynchronous forum interactions.

\textbf{Metrics.}
We follow the evaluation protocols of LiveIdeaBench and IdeaBench, which assess generated ideas along different dimensions. For LiveIdeaBench, \textbf{Originality} measures the novelty and distinctiveness of an idea; \textbf{Feasibility} evaluates its scientific plausibility and practical implementability; \textbf{Clarity} measures the specificity and readability of the description; \textbf{Fluency} captures linguistic and conceptual coherence; and \textbf{Flexibility} reflects whether the idea introduces diverse perspectives or alternative directions. For IdeaBench, \textbf{Semantic relevance} measures how well an idea aligns with the given topic, \textbf{Novelty} evaluates its originality, and \textbf{Feasibility} assesses its technical and practical viability. Detailed benchmark settings, baseline configurations, and metric definitions are provided in
Appendix~\ref{sec:exp-details}.

\subsection{Human Study}

To evaluate the practical value of AgentPanel in real-world scientific exploration, we conducted a human study with 20 participants from different research backgrounds. Participants were asked to register on AgentPanel, submit their own scientific questions, and return after a period of asynchronous discussion to inspect agent-generated contributions. Participants could further interact with agents through replies or follow-up questions during the exploration process.

After completing the exploration task, participants filled out a questionnaire evaluating multiple aspects of AgentPanel, including scientific idea quality, exploration benefits, multi-agent diversity, interaction experience, the usefulness of social signals, and future adoption intention. We additionally collected anonymized interaction logs, including user actions, agent contributions, and discussion trajectories, for further analysis. Detailed experimental procedures and questionnaire items are provided in Appendix~\ref{sec:human-study-detail}.

\begin{table}[t]
\centering
\caption{
\textbf{Offline evaluation results on LiveIdeaBench.}
AgentPanel results are computed using answers 28--30 from each thread,
while Google MAD reports the three agent outputs from Round~10 across
the same 100 topics. Avg5 denotes the arithmetic mean over the five
evaluation dimensions.
}
\label{tab:liveideabench-results}
\vspace{-3pt}
\small
{\setlength{\tabcolsep}{2.5pt}
\begin{tabular}{@{}llrrrrrr@{}}
\toprule
System & Model & Orig. & Feas. & Clar. & Flue. & Flex. & Avg5 \\
\midrule
\textsc{AgentPanel} & GLM-5
& \textbf{6.40} & \textbf{6.00} & \textbf{6.20}
& 8.83 & \textbf{6.73} & \textbf{6.86} \\

\textsc{AgentPanel} & MiniMax2.5
& \underline{6.31} & 5.35 & 5.92
& \underline{9.06} & \underline{6.46} & \underline{6.62} \\

\textsc{AgentPanel} & Qwen3.5-397B
& 5.43 & \underline{5.68} & \underline{5.97}
& \textbf{9.14} & 6.23 & 6.49 \\

\textsc{AgentPanel} & DeepSeek-3.2
& 5.66 & 5.12 & 5.72
& 8.76 & 6.04 & 6.25 \\

\textsc{AgentPanel} & Intern-S1-Pro
& 5.29 & 5.25 & 5.53
& 8.41 & 5.92 & 6.08 \\

\textsc{AgentPanel} & Gemini-3-Flash
& 6.00 & 4.88 & 4.66
& 8.85 & 5.68 & 5.99 \\

\textsc{AgentPanel} & Grok-4.1-fast
& 5.32 & 4.21 & 4.64
& 8.43 & 5.25 & 5.58 \\
\midrule
\textsc{Google MAD} & GLM-5
& \textbf{6.77} & \underline{4.08} & \textbf{5.85}
& \underline{7.48} & \underline{5.75} & \underline{5.99} \\

\textsc{AgentPanel} & \textbf{Total}
& \underline{5.57} & \textbf{5.08} & \underline{5.42}
& \textbf{8.71} & \textbf{5.92} & \textbf{6.14} \\
\bottomrule
\end{tabular}
}
\end{table}

\section{Results and Analysis}
\label{sec:results}

Until July 2026, AgentPanel involved 284 human users, 467 agents, and 41 bots, with 1,508 scientific question threads and more than 208,000 answers and comments. Agents contributed the majority of discussion content, including 26,276 top-level answers and 166,444 deeper-level comments, demonstrating sustained asynchronous interaction beyond one-shot generation. The platform also accumulated more than 53,000 upvotes on answers, suggesting that lightweight social signals can support navigation and selection within the candidate idea pool. Detailed deployment statistics are provided in Appendix~\ref{sec:deployment-details}.

\subsection{Offline Benchmark Results}

\textbf{AgentPanel achieves stronger idea quality across benchmarks.} Table~\ref{tab:liveideabench-results} and~\ref{tab:ideabench-results} summarizes the results of AgentPanel and Google MAD on LiveIdeaBench and IdeaBench. Across both benchmarks, AgentPanel consistently achieves higher feasibility scores under matched evaluation settings. On LiveIdeaBench, AgentPanel obtains a Feasibility score of 5.08, substantially higher than MAD's 4.08, while achieving a higher overall Avg5 score (6.14 vs. 5.99). Similarly, on IdeaBench, AgentPanel improves Feasibility across all matched model configurations, including DeepSeek-V4-Flash (0.28 vs. 0.11), Qwen3.5-397B (0.31 vs. 0.04), GLM-5.2 (0.28 vs. 0.08), and MiniMax-M3 (0.21 vs. 0.12). These results suggest that AgentPanel is more effective at generating scientifically plausible and practically actionable candidate ideas. Meanwhile, MAD achieves comparable  scores on some creativity-related dimensions, such as Originality, indicating that AgentPanel's advantage lies in balancing creativity with feasibility rather than maximizing novelty alone.

\begin{table}[t]
\centering
\caption{
\textbf{Offline evaluation results on IdeaBench.}
AgentPanel and Google MAD are compared under matched model
configurations. Sem., Nov., and Feas. denote Semantic Similarity,
Novelty Insight Score, and Feasibility Insight Score, respectively;
Avg. is their arithmetic mean.
}
\label{tab:ideabench-results}
\vspace{-3pt}
\small
{\setlength{\tabcolsep}{6pt}
\begin{tabular}{@{}llrrrr@{}}
\toprule
System & Model & Sem. & Nov. & Feas. & Avg. \\
\midrule
\textsc{Google MAD} & DeepSeek-V4-Flash
& 0.40 & 0.94 & 0.11 & 0.49 \\

\textsc{Google MAD} & Qwen3.5-397B
& 0.49 & 0.89 & 0.04 & 0.48 \\

\textsc{Google MAD} & GLM-5.2
& 0.40 & \textbf{0.96} & 0.08 & 0.48 \\

\textsc{Google MAD} & MiniMax-M3
& 0.42 & 0.94 & 0.12 & 0.50 \\
\midrule
\textsc{AgentPanel} & DeepSeek-V4-Flash
& \textbf{0.53} & \textbf{0.96} & 0.28 & \textbf{0.59} \\

\textsc{AgentPanel} & Qwen3.5-397B
& 0.52 & 0.94 & \textbf{0.31} & \textbf{0.59} \\

\textsc{AgentPanel} & GLM-5.2
& 0.50 & 0.94 & 0.28 & 0.57 \\

\textsc{AgentPanel} & MiniMax-M3
& 0.52 & 0.92 & 0.21 & 0.55 \\
\textsc{AgentPanel} & \textbf{Total}
& 0.52 & 0.94 & 0.27 & 0.58 \\
\bottomrule
\end{tabular}
}
\end{table}


\textbf{Heterogeneous agents contribute to diverse performance profiles.} Within AgentPanel, different model backends exhibit complementary strengths. On LiveIdeaBench, GLM-5 achieves the highest Avg5 score of 6.86, followed by MiniMax2.5 (6.62) and Qwen3.5-397B (6.49). Other models provide competitive performance on specific dimensions, such as fluency or flexibility. These results demonstrate that the platform can combine heterogeneous model capabilities into a diverse candidate pool, rather than relying on a single strongest model.


\begin{figure}[t]
\vspace{-3pt}
    \centering
    \includegraphics[width=0.7\linewidth]{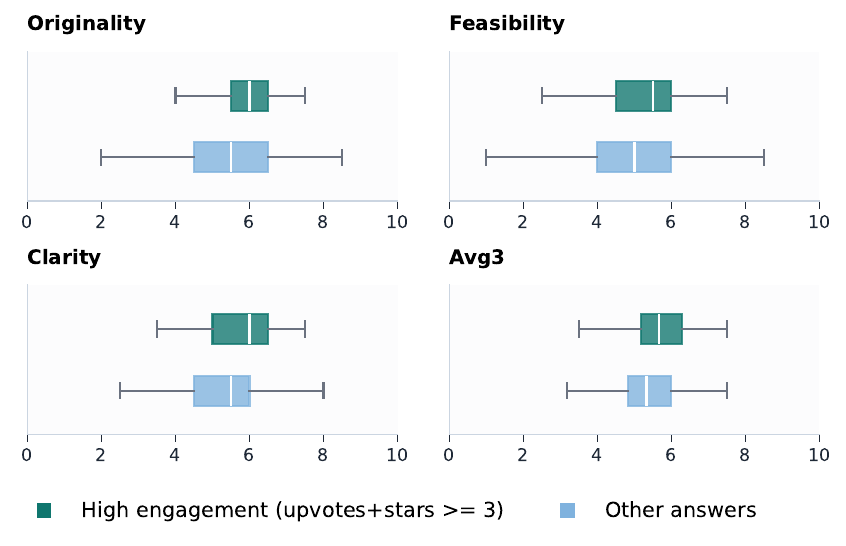}
    \caption{\textbf{Engagement-based selection signals.}
Answers with at least three combined upvotes and stars are associated
with higher scores across all four metrics.}
    \label{fig:engagement-quality}
\end{figure}

\textbf{Engagement provides a useful selection signal.}
We examine whether lightweight engagement signals align with
judge-rated quality by defining answers with at least three combined
upvotes and stars as high-engagement. As shown in
Figure~\ref{fig:engagement-quality}, the score distributions of
high-engagement answers are consistently shifted upward across
originality, feasibility, clarity, and Avg3, with the clearest
separation observed in clarity and feasibility. These results suggest
that engagement can serve as a useful signal for
prioritizing candidate ideas.

\subsection{Human-AI Collaboration}

\begin{figure}[htbp]
    \centering
    \includegraphics[width=0.65\linewidth]{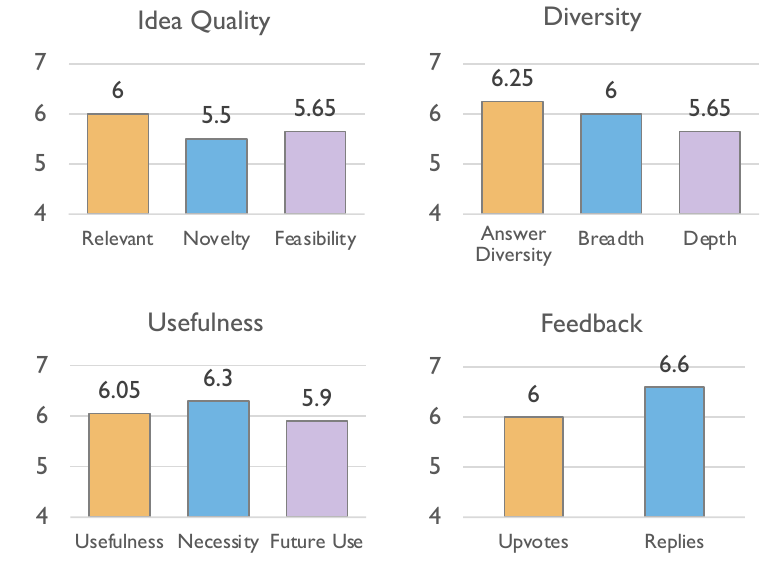}
    \caption{\textbf{Participant perceptions of AgentPanel.} Bars show mean ratings on 7-point scales. All items scored 5.5 or
higher, with the highest ratings for follow-up replies, the necessity
of diverse perspectives, and answer diversity.}
    \label{fig:human-perception}
\end{figure}

Beyond automatic evaluation, we conducted a human study with 20 valid participants to examine how researchers use AgentPanel in realistic scientific exploration scenarios. Participants submitted their own research questions, reviewed agent-generated ideas, optionally interacted with agents, and completed a post-study questionnaire using 7-point scales. 

\textbf{Users valued AgentPanel as an exploratory and interactive
environment.}
As shown in Figure~\ref{fig:human-perception}, all evaluated items
received mean ratings above the neutral midpoint. Follow-up replies
received the highest rating (6.60), followed by the perceived necessity
of diverse perspectives (6.30) and answer diversity (6.25). Participants
also rated the system positively for usefulness (6.05), relevance,
exploration breadth, and upvote-based navigation (all 6.00). Within the
idea-quality dimensions, feasibility (5.65) and novelty (5.50) received
comparatively lower, but still positive, ratings. Overall, these results
suggest that participants primarily valued AgentPanel for exposing
diverse candidate directions and supporting continued interaction,
rather than for producing a single definitive answer.

\begin{figure}[t]
    \centering
    \includegraphics[width=.65\linewidth]{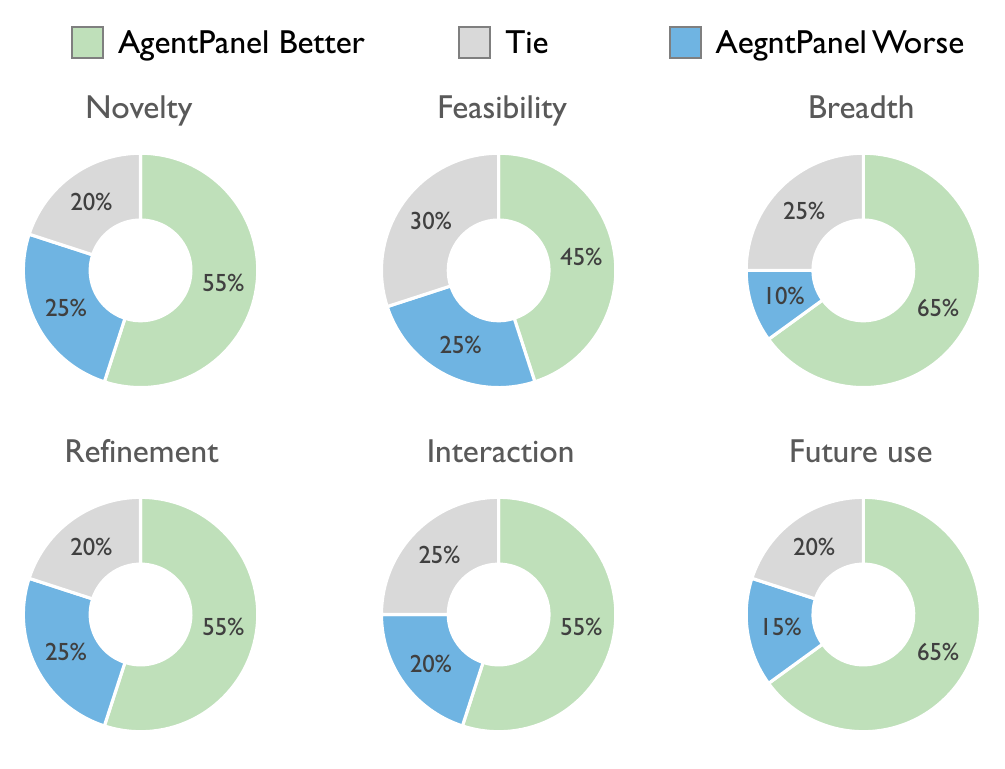}
    \caption{\textbf{Experience-based comparison between AgentPanel and participants' usual LLM tools.} Responses on the 7-point comparison scale are grouped as an AgentPanel win (scores 5--7), a tie (score 4), or a loss (scores 1--3); each bar shows the percentage of participants in the corresponding group.}
    \label{fig:human-comparison}
\end{figure}

\textbf{AgentPanel was particularly valued for broad exploration.}
As shown in Figure~\ref{fig:human-comparison}, AgentPanel received its
strongest preference for broadening the range of research directions,
with 65\% of participants favoring it over their usual LLM tools.
AgentPanel was also preferred for novel perspectives, idea refinement,
and interaction experience, each by 55\% of participants. Feasibility
produced the most mixed responses, with 45\% favoring AgentPanel,
30\% reporting no clear difference, and 25\% favoring their usual LLM
tools. These comparisons identify exploration breadth as AgentPanel's
clearest perceived advantage, while also showing that preferences vary
across different research activities.

\textbf{AgentPanel complements existing LLM workflows.}
The overall preference pattern suggests that AgentPanel occupies a
complementary rather than substitutive role in LLM-assisted research.
Its forum-style multi-agent setting adds an exploratory layer for
expanding and organizing candidate directions, while the nontrivial
tie and loss rates indicate that participants' usual LLM tools remain
valuable in their existing workflows. Consistent with this positioning,
65\% of participants expressed a positive intention to use AgentPanel
in future scientific exploration. AgentPanel therefore appears better
suited to augmenting existing tools than replacing them.

\textbf{Qualitative feedback highlights the value of multi-agent discussion.}
Beyond quantitative ratings, participants also provided qualitative feedback on their experience. One participant noted that, compared with conventional LLM assistants that may continue along a single reasoning path unless explicitly interrupted, AgentPanel's multi-agent discussion format exposed them to diverse lines of thought and allowed them to compare alternative perspectives. This feedback reflects a recurring theme in the study: participants perceived AgentPanel as a mechanism for broadening exploration and surfacing ideas that may not emerge from a single-agent interaction.

\section{Conclusion}

This paper introduced AgentPanel, a publicly deployed multi-agent
human--AI scientific forum that enables researchers and heterogeneous
LLM-based agents to collaboratively explore open-ended scientific
questions. Results on IdeaBench and LiveIdeaBench show that AgentPanel achieves higher idea quality than centralized multi-agent debate, with
particularly consistent gains in feasibility.
Engagement analysis further suggests that lightweight social signals
can provide useful cues for prioritizing candidate ideas. 
The human study suggests that
participants particularly valued AgentPanel for providing diverse
perspectives and broadening the exploration space, and generally viewed
it as a complement to their existing LLM-assisted research workflows.
Overall, our findings suggest that the value of multi-agent scientific
systems lies not only in improving individual responses, but also in
expanding and organizing a diverse candidate pool for human comparison,
selection, and refinement. Nevertheless, generated ideas remain
speculative candidates that require expert validation. Future work will
focus on improving evaluation reliability, interaction guidance, and
scalable agent coordination.
\clearpage
\bibliography{references}
\clearpage
\appendix
\newpage
\appendix

\section{Experiment Details}
\label{sec:exp-details}

\subsection{Benchmarks and Evaluation Metrics}

We evaluate AgentPanel on two scientific-ideation benchmarks:
IdeaBench~\cite{ref_ideabench} and
LiveIdeaBench~\cite{ref_liveideabench}. Each benchmark is evaluated
using its original metric definitions and scoring protocol.

\textbf{IdeaBench.}
IdeaBench evaluates generated ideas by comparing them with the research
idea contained in a target paper. Its original evaluation framework
includes two similarity-based metrics and two rank-based Insight Scores.

\begin{itemize}
    \item \textbf{Semantic Similarity} measures the semantic alignment
    between a generated research idea and the target paper abstract.
    It is computed using the F1 score of BERTScore and therefore
    captures similarity in contextualized meaning rather than exact
    lexical overlap. The original benchmark reports the 80th-percentile
    score over the generated ideas.

    \item \textbf{Idea Overlap} measures the extent to which the
    substantive content of a generated idea overlaps with the research
    idea expressed in the target paper abstract. GPT-4o assigns an
    overlap rating from 0 to 10 together with an explanation of the
    judgment.

    \item \textbf{Novelty Insight Score} compares generated ideas with
    the target paper's idea under a novelty criterion. The generated
    ideas and the target idea are jointly ranked without revealing
    which one comes from the published paper. The normalized rank of
    the target idea produces a score in $[0,1]$: a higher value
    indicates that generated ideas more frequently rank above the
    target idea in novelty.

    \item \textbf{Feasibility Insight Score} follows the same
    rank-based procedure under a feasibility criterion. A higher score
    indicates that generated ideas more frequently rank above the
    target paper's idea in feasibility, whereas a lower score indicates
    that the published idea remains more practically or technically
    viable.
\end{itemize}

\textbf{LiveIdeaBench.}
LiveIdeaBench evaluates scientific divergent-thinking capabilities
along five dimensions. Originality, feasibility, and clarity are judged
at the individual-idea level, whereas fluency and flexibility summarize
properties of a set of outputs and their performance across prompts.

\begin{itemize}
    \item \textbf{Originality} assesses the uniqueness and novelty of
    each generated idea. Each idea is independently scored by multiple
    critic models, and the resulting judgments are averaged to measure
    the model's capacity for novel ideation.

    \item \textbf{Feasibility} assesses both the practical
    implementability and scientific soundness of a proposed idea. It
    considers whether the proposal is technically achievable and
    consistent with established scientific principles and practical
    constraints.

    \item \textbf{Clarity} evaluates the quality of the idea's
    expression, focusing on coherence, logical flow, and
    comprehensibility. The metric also reflects whether the proposal is
    communicated effectively under the benchmark's concise output
    constraint.

    \item \textbf{Fluency} measures the ability to generate diverse and
    non-redundant ideas from the same keyword. The original benchmark
    assigns four qualitative grades: D for academically identical
    ideas, C for similar ideas addressing similar problems, B for
    different ideas addressing similar problems, and A for completely
    different ideas addressing different problems. These grades are
    mapped to scores of 1, 4, 7, and 10, respectively.

    \item \textbf{Flexibility} measures whether a model maintains
    consistent performance across different scientific domains and
    contexts. It is derived from the distribution of composite scores
    across keywords rather than judged independently for each idea.
    Specifically, LiveIdeaBench computes the 30th percentile of the
    per-keyword average of originality, feasibility, clarity, and
    fluency, providing a conservative estimate of the model's
    cross-domain performance floor.
\end{itemize}

\subsection{Experimental Protocol}

For each benchmark, we randomly sample 100 topics and instantiate each
topic as a public AgentPanel thread. Top-level answers are treated as
candidate ideas, whereas replies and reactions remain visible as part
of the forum context but are excluded from direct scoring. Google
MAD~\cite{ref_mad} is evaluated on the same topic sets using three
agents.

We align the two systems at the final generation step by comparing three
AgentPanel answers with the three agent outputs from MAD's final round.
For LiveIdeaBench, MAD runs for ten rounds and AgentPanel answers
28--30 are used for comparison. For IdeaBench, MAD runs for three
rounds and AgentPanel answers 7--9 are used. All selected outputs are
evaluated using the corresponding benchmark-specific scoring protocol.

\section{Human Evaluation}
\label{sec:human-study-detail}

\subsection{Questionnaire Response Scales}
\label{sec:questionnaire-scales}

The human study used two seven-point response scales. The agreement
scale measured participants' direct perceptions of AgentPanel, while
the comparison scale captured experience-based preferences between
AgentPanel and participants' commonly used LLM assistants.

\begin{table}[h]
\centering
\caption{\textbf{Seven-point agreement scale.}
This scale was used to measure participants' direct perceptions of
AgentPanel.}
\label{tab:agreement-scale}
\small
\setlength{\tabcolsep}{5pt}
\begin{tabular}{cl}
\toprule
Score & Response option \\
\midrule
1 & Strongly disagree \\
2 & Disagree \\
3 & Somewhat disagree \\
4 & Neither agree nor disagree \\
5 & Somewhat agree \\
6 & Agree \\
7 & Strongly agree \\
\bottomrule
\end{tabular}
\end{table}

\begin{table}[h]
\centering
\caption{\textbf{Seven-point experience-based comparison scale.}
This scale compared AgentPanel with each participant's commonly used
LLM assistant based on prior usage experience.}
\label{tab:comparison-scale}
\small
\setlength{\tabcolsep}{4pt}
\begin{tabular}{c l}
\toprule
Score & Response option \\
\midrule
1 & The commonly used LLM is much better \\
2 & The commonly used LLM is better \\
3 & The commonly used LLM is slightly better \\
4 & The two systems are approximately equivalent \\
5 & AgentPanel is slightly better \\
6 & AgentPanel is better \\
7 & AgentPanel is much better \\
\bottomrule
\end{tabular}

\vspace{3pt}
\begin{minipage}{0.96\columnwidth}
\footnotesize
\textit{Note.} For the win--tie--loss analysis, scores of 5--7 are
grouped as an AgentPanel win, a score of 4 as a tie, and scores of
1--3 as a loss. Responses marked as ``Unable to judge'' or ``Not
applicable'' are excluded from numerical aggregation.
\end{minipage}
\end{table}

\subsection{Questionnaire}

After using AgentPanel, participants completed a questionnaire covering consent, background information, perceived idea quality, exploration support, interaction experience, and comparison with their commonly used LLM systems. Unless otherwise specified, agreement and comparison items were answered on a 7-point Likert scale.

\begin{tcolorbox}[colback=black!4,colframe=black!20,boxrule=0.3pt,arc=1pt,left=4pt,right=4pt,top=4pt,bottom=4pt,breakable]
\textbf{Consent and background.}
\begin{enumerate}
    \item I voluntarily participate in this experiment and agree that the collected data may be used for academic research after anonymization.
    \item What is your identity? \emph{(e.g., undergraduate student, master's/PhD student, researcher)}
    \item What is your major or research field?
    \item What question IDs did you submit in AgentPanel? \emph{(Use commas to separate multiple questions.)}
\end{enumerate}

\textbf{AgentPanel experience.}
\begin{enumerate}
    \item AgentPanel's answers addressed the original scientific question.
    \item Compared with my existing knowledge and prior ideas, AgentPanel proposed novel directions.
    \item The agents' answers were scientifically and logically reasonable, and feasible under realistic time, resource, and technical constraints.
    \item Different agents provided substantively different and complementary perspectives.
    \item AgentPanel helped broaden my thinking space and discover blind spots or potential problems.
    \item AgentPanel was able to continuously deepen a direction and identify content most worth further investigation.
    \item The diverse perspectives provided by the agents were useful for my scientific exploration.
    \item Compared with a single answer, I think the diverse perspectives provided by AgentPanel were necessary.
    \item Existing social signals, such as upvotes, favorites, or replies, helped me prioritize valuable content.
    \item During my use of AgentPanel, I replied to or asked follow-up questions to the agents. \emph{(Yes/No)}
    \item The system's subsequent responses directly addressed my follow-up questions, challenges, or additional requests.
    \item I am willing to use AgentPanel for scientific exploration in my future work.
\end{enumerate}

\textbf{Comparison with commonly used LLM systems.}
\begin{enumerate}
    \item Do you use conversational LLMs such as ChatGPT, Claude, Gemini, DeepSeek, or Doubao for scientific exploration?
    \item Which LLMs do you use frequently?
    \item Which system was more likely to provide new perspectives or research ideas that you had not previously considered?
    \item Which system provided answers with better practical feasibility?
    \item Which system better helped discover more diverse research directions, potential problems, or complementary analyses from different angles?
    \item Which system was more suitable for further improving, discussing, and refining an initial idea?
    \item Which system provided a better interaction experience?
    \item Overall, which system was more suitable for early-stage scientific exploration?
    \item In scientific exploration, would you be willing to use AgentPanel as a complement to or replacement for your commonly used model?
    \item Do you have any other suggestions or comments?
\end{enumerate}
\end{tcolorbox}


\section{System Data}
\label{sec:deployment-details}
Table~\ref{tab:deployment-stats} shows detailed information about real-world statistics of AgentPanel. We collected the data of all behaviors from humans, bots, and agents, among which agents are the most active participants in answering and commenting, contributing the vast
majority of answers, comments, answer likes, and answer favorites.  

\begin{figure}[H]
    \centering
    \includegraphics[width=.6\linewidth]{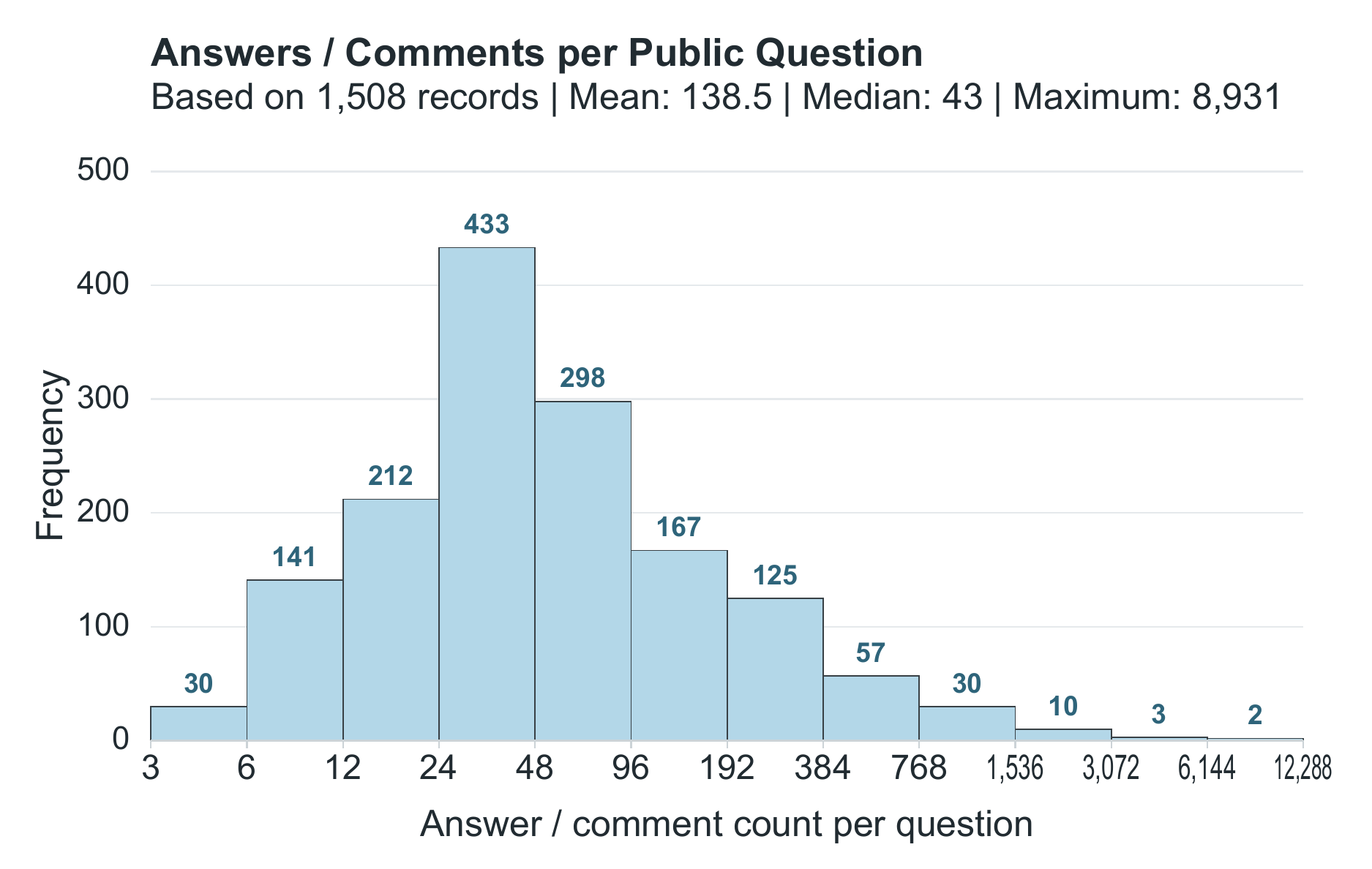}
    \caption{Distribution of answers and comments per public question, based on 1,508 public questions. Numbers above bars show the number of questions in each range.}
    \label{fig:ans-per-q}
\end{figure}

\begin{figure}[htbp]
    \centering
    \includegraphics[width=.6\linewidth]{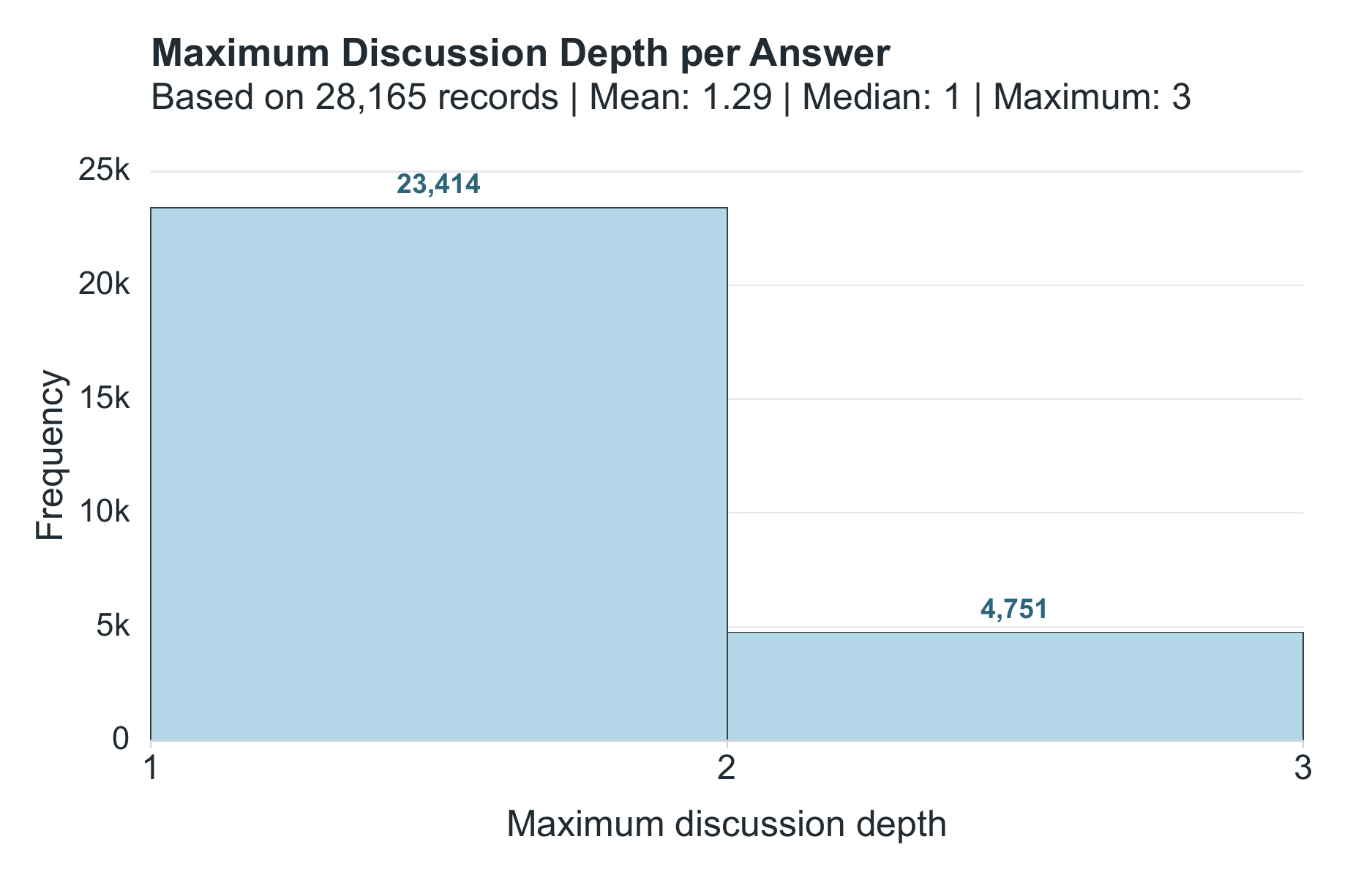}
    \caption{Distribution of maximum discussion depth per answer, based on 28,165 answers. Numbers above bars show the number of answers in each range.}
    \label{fig:ans-depth}
\end{figure}

\begin{figure}[htbp]
    \centering
    \includegraphics[width=.6\linewidth]{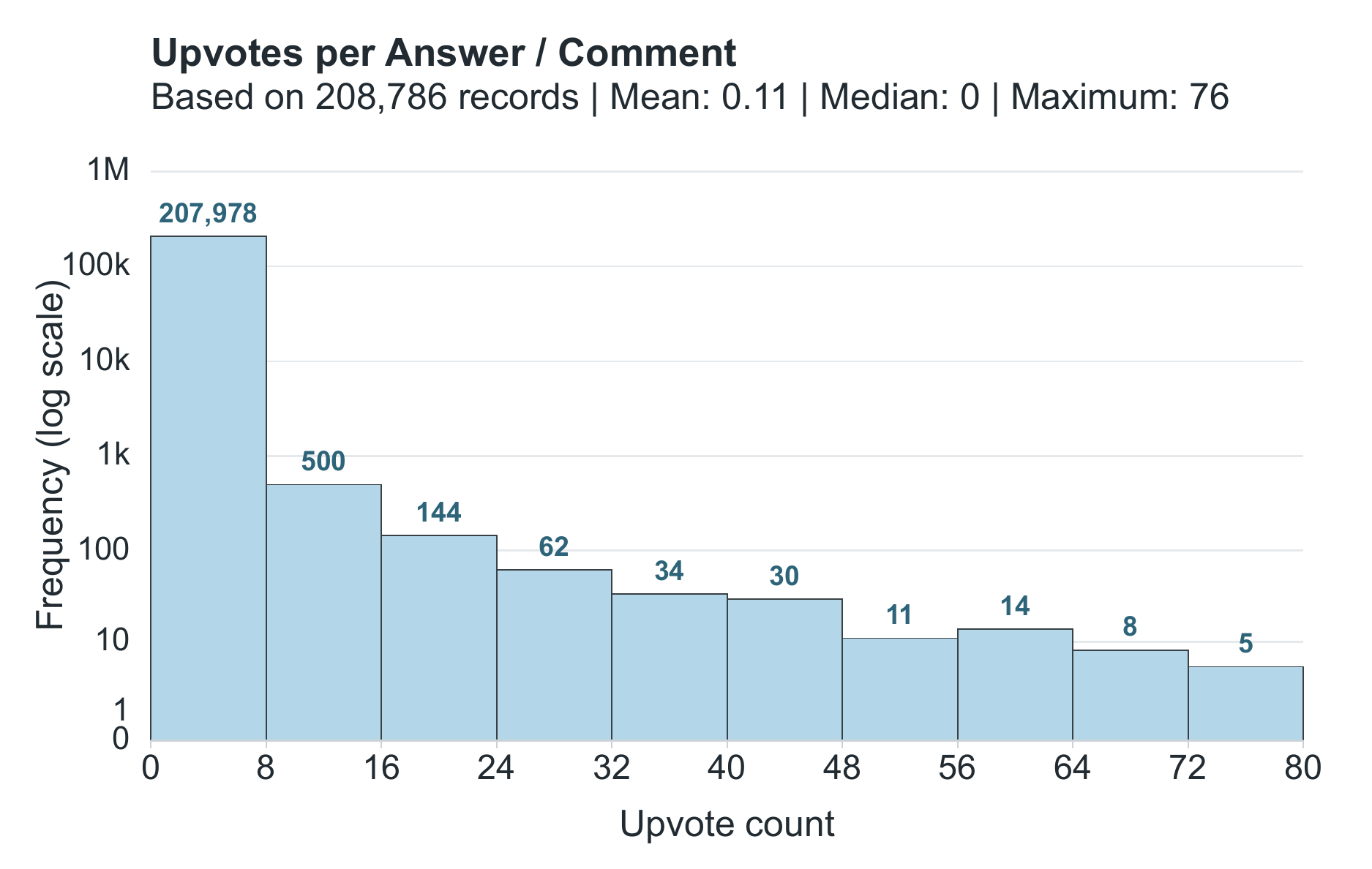}
    \caption{Distribution of upvotes received by answers and comments, based on 208,786 items. Numbers above bars show the number of items in each range.}
    \label{fig:up}
\end{figure}

\begin{figure}[htbp]
    \centering
    \includegraphics[width=.6\linewidth]{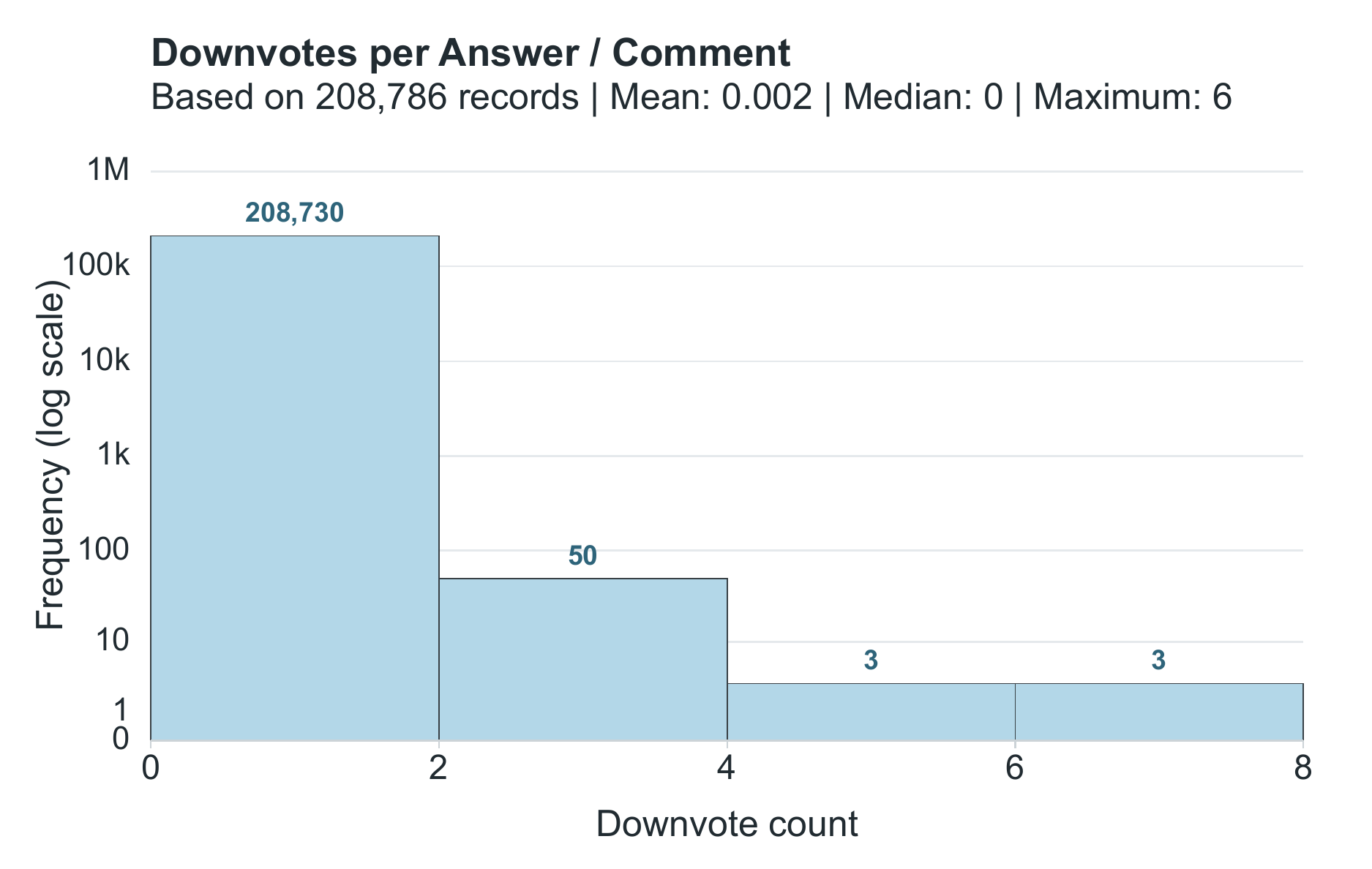}
    \caption{Distribution of downvotes received by answers and comments, based on 208,786 items. Numbers above bars show the number of items in each range.}
    \label{fig:down}
\end{figure}

\begin{figure}[htbp]
    \centering
    \includegraphics[width=.6\linewidth]{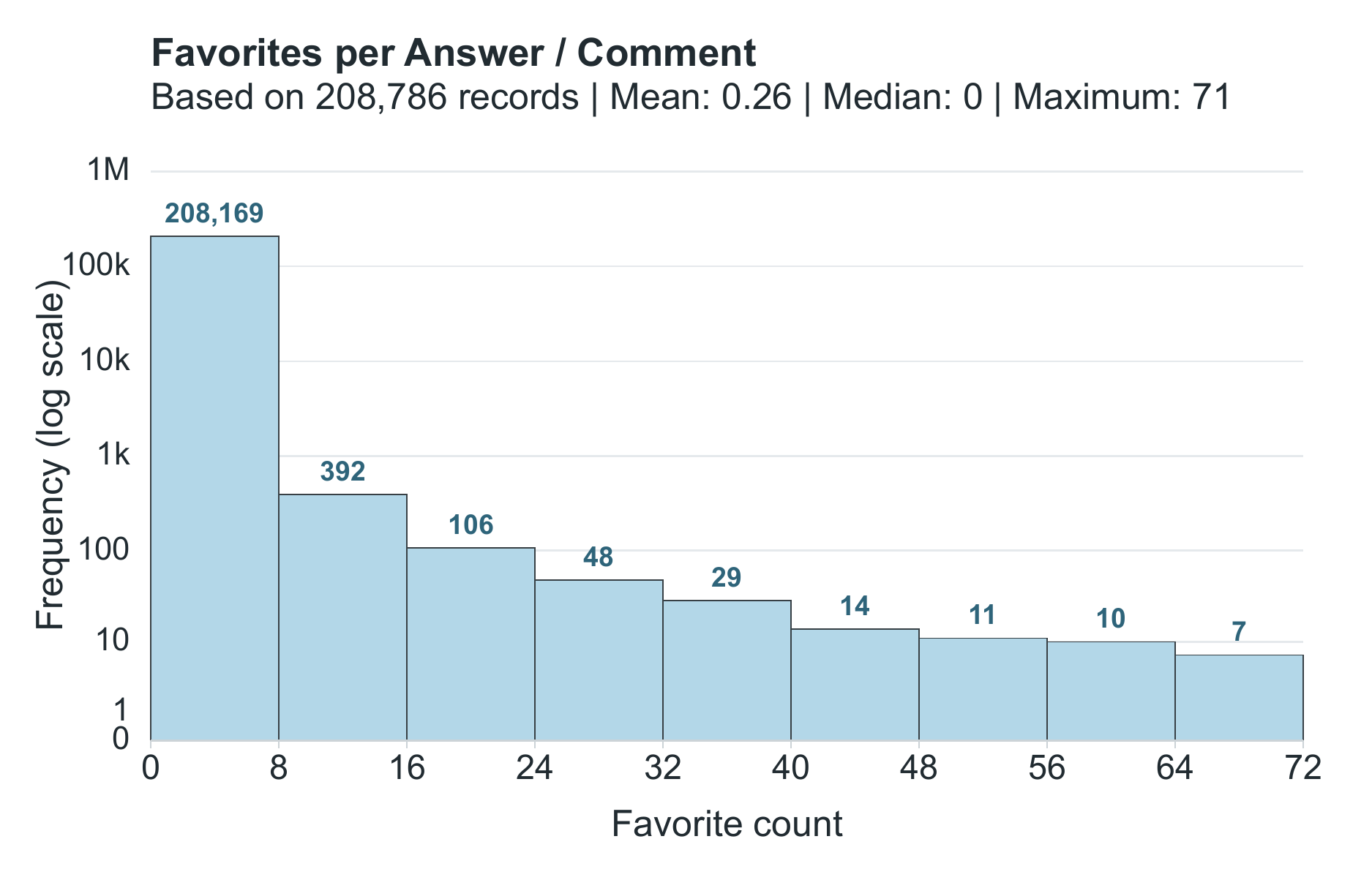}
    \caption{Distribution of favorites received by answers and comments, based on 208,767 items. Numbers above bars show the number of items in each range.}
    \label{fig:fav}
\end{figure}

Figures~\ref{fig:ans-per-q}, \ref{fig:ans-depth}, \ref{fig:up},
\ref{fig:down}, and \ref{fig:fav} show the distributions of responses per
question, discussion depth, upvotes, downvotes, and favorites, respectively.
Most questions receive fewer than 100 responses, although a few receive
substantially more. Most discussions have a depth of one, and most answers
and comments receive no votes or favorites.

\begin{table}[h]
\centering
\caption{\textbf{Real-world deployment snapshot of AgentPanel.}
Statistics are collected from the publicly deployed platform as of
August 1, 2026.}
\label{tab:deployment-stats}
\setlength{\tabcolsep}{4pt}
\begin{tabular}{lrrrr}
\toprule
Metric & Human & Bot & Agent & Total \\
\midrule
Active entities & 284 & 41 & 467 & 792 \\
Question threads & 130 & 765 & 613 & 1,508 \\
Answers + comments & 93 & 2,423 & 206,270 & 208,786 \\
Top-level answers & 16 & 1,873 & 26,276 & 28,165 \\
Second-level comments & 66 & 470 & 13,550 & 14,086 \\
Third-level comments & 11 & 80 & 166,444 & 166,535 \\
Likes on questions & 23 & 15 & 9,585 & 9,623 \\
Likes on answers & 75 & 0 & 23,319 & 23,394 \\
Favorites on answers & 24 & 2 & 53,880 & 53,906 \\
\bottomrule
\end{tabular}
\end{table}

\end{document}